\documentclass[letterpaper, 10 pt, conference]{ieeeconf}  

\IEEEoverridecommandlockouts                              

\usepackage[noadjust]{cite}

\usepackage{graphicx}
\usepackage{amsmath} 
\usepackage{amssymb}  
\usepackage{balance}
\usepackage[T1]{fontenc}
\usepackage{tikz}

\usepackage{hyperref}
\usepackage[bottom]{footmisc} 

\usepackage{caption}
\usepackage{subcaption}

\usepackage[ruled, linesnumbered, vlined]{algorithm2e}
\SetAlFnt{\small}
\SetAlCapFnt{\small}
\SetAlCapNameFnt{\small}

\usepackage{upgreek}
\usepackage{bm}

\definecolor{colorfor1}{HTML}{6da4f3}
\definecolor{colorfor2}{HTML}{8ec18b}
\definecolor{colorfor3}{HTML}{ea6e68}

\definecolor{significant_001}{HTML}{68c79c}
\definecolor{significant_01}{HTML}{4f9a90}
\definecolor{significant_05}{HTML}{fa8c63}

\DeclareMathOperator*{\argmin}{arg\,min}

\title{\LARGE \bf Hey Robot! Personalizing Robot Navigation through Model Predictive Control with a Large Language Model}

\author{Diego Martinez-Baselga$^{1}$, Oscar de Groot$^{2}$, Luzia Knoedler$^{2}$,\\Javier Alonso-Mora$^{2}$, Luis Riazuelo$^{1}$, Luis Montano$^{1}$
 \thanks{
 $^{1}$ The authors are with RoPeRt Group, DIIS-I3A, University of Zaragoza, 50018 Zaragoza, Spain. (\texttt{\small \{diegomartinez, riazuelo, montano\}@unizar.es}).}
\thanks{
 $^{2}$The authors are with the Dept. of Cognitive Robotics, TU Delft, 2628 CD Delft, The Netherlands. (\texttt{\small \{o.m.degroot, l.knoedler, j.alonsomora\}@tudelft.nl}).}%
\thanks{This work was supported by the Ministry of Science, Innovation and Universities of Spain through the research project MCIN/AEI/10.13039/501100011033/FEDER-UE, from the European Union’s Horizon 2020 research and innovation programme under grant agreement No. 101017008, and from the European Union (ERC, INTERACT, 101041863). Views and opinions expressed are however those of the author(s) only and do not necessarily reflect those of the European Union or the European Research Council Executive Agency. Neither the European Union nor the granting authority can be held responsible for them.}%
}
\begin{document}

\maketitle
\thispagestyle{empty}
\pagestyle{empty}

\begin{abstract}
Robot navigation methods allow mobile robots to operate in applications such as warehouses or hospitals. While the environment in which the robot operates imposes requirements on its navigation behavior, most existing methods do not allow the end-user to configure the robot's behavior and priorities, possibly leading to undesirable behavior (e.g., fast driving in a hospital). We propose a novel approach to adapt robot motion behavior based on natural language instructions provided by the end-user. Our zero-shot method uses an existing Visual Language Model to interpret a user text query or an image of the environment. This information is used to generate the cost function and reconfigure the parameters of a Model Predictive Controller, translating the user's instruction to the robot's motion behavior. This allows our method to safely and effectively navigate in dynamic and challenging environments. We extensively evaluate our method's individual components and demonstrate the effectiveness of our method on a ground robot in simulation and real-world experiments, and across a variety of environments and user specifications.

\end{abstract}
\section{Introduction}
As mobile robots increasingly operate around humans, ensuring social and safe interactions becomes essential. However, robot's behavior requirements depend on the environment, its crowd density, and the specific robot tasks. For instance, social aspects are important requirements for elderly care while efficiency is crucial in warehouse collaboration. A mobile robot should navigate safely in all of these scenarios, while its required tasks and behavior differ.

Recent works on socially aware navigation conclude that robots need to accommodate to human behavior and understand human intentions~\cite{mavrogiannis2023core, francis2023principles}, as not satisfying user preferences leads to negative experiences and frustration~\cite{kruse2013human}. Behaviors can generally be configured by adapting the cost function of the motion controller. 
However, this requires in-depth knowledge of the system and is seldom manageable by the end user. On the other hand, designing behaviors for all types of scenarios is unattainable. Requirements may even change over time as humans become more accustomed to the robot. 
Instead, the end-user should be able to modify the robot's behavior without requiring expert knowledge.


\begin{figure}[t]
    \centering
    \includegraphics[width=\linewidth]{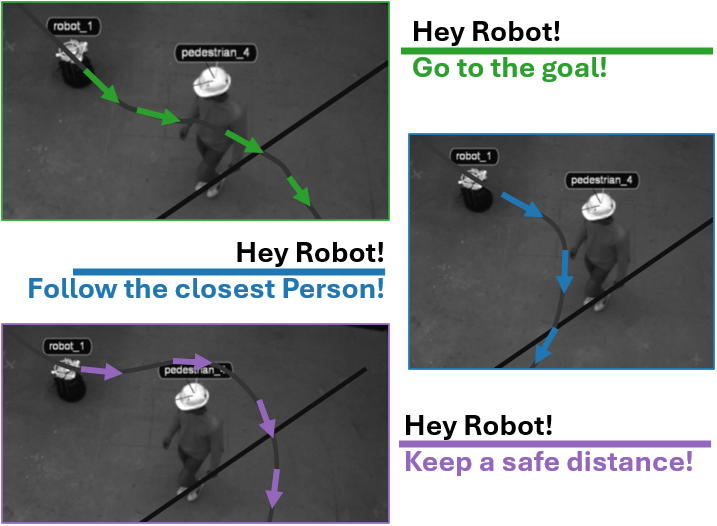}
    \caption{A user asks the robot for a motion behavior, which is fulfilled in real-time.}
    \label{fig:teaser}
\end{figure}

Large Language Models (LLM) and Visual Language Models (VLM) like GPT-4~\cite{achiam2023gpt}, LLaMa~\cite{touvron2023llama} or Gemini~\cite{reid2024gemini} have emerged as powerful tools to process 
natural language and images and are capable of complex tasks. 
They pose an opportunity to interpret user requirements and requests directly from natural language. 
Additionally, while these methods are not trained on the specific end-user environment, they have a general understanding of real-world scenarios that can be leveraged to inform navigation behavior. 


In this work, we propose an architecture composed of existing pre-trained LLM assistants and a Model Predictive Control (MPC)~\cite{de2024topology} planner to navigate challenging dynamic scenarios while accommodating user specifications, as shown in Fig.~\ref{fig:teaser}. Our key idea is to use the cost function of the MPC to assess and modify the capabilities of the planner in order to satisfy the user's query. Our method leverages LLMs to interpret and generate the code implementing the cost function, translating natural language instructions (e.g., "go slower", "keep more distance to people", "navigate as if you were in a hospital", "follow the path", "follow the closest human", etc.) to the desired navigation behavior. 
By adapting only the cost function of the MPC, we allow the user to change the robot behavior on the fly, while maintaining planner safety via constraints. As an additional feature, we equip the robot with a camera to allow it to perceive the environment independently, enabling direct adaptation based on visual input if requested. 
We extensively evaluate the components of the method and the resulting navigation stack both in simulation and in hardware experiments. To this end, our contribution is three-fold:
\begin{itemize}
    \item A novel zero-shot method to generate and tune cost functions for an MPC planner from human natural language and, if requested, images taken from a robot onboard camera.
    \item A novel MPC motion planner framework for dynamic environments that may change interactively and on the fly its navigation behavior with user instructions.
    \item Extensive simulation and ground robot experiments that assess the performance of the proposed planner given different specified requirements.
\end{itemize}
Our implementation, that will be released upon acceptance, uses ROS, C++, Python and Acados~\cite{Verschueren2021acados} to enable real-time interactions.
\section{Related Work}

\textbf{Navigation in human-centered environments.} Mobile robot navigation in environments shared with humans is often framed as a collision avoidance problem with dynamic obstacles. 
It can be approached with purely reactive methods, such as velocity obstacle-based approaches~\cite{alonso2018cooperative, martinez2024avocado} and the social force model~\cite{helbing1995social}. 
Another important body of work addresses multi-agent interactions from a game-theoretic perspective~\cite{peters2024contingency,mehr2023maximum}, 
but they present a considerable computational challenge. Reinforcement learning-based methods~\cite{chen2019crowd, everett2021collision, martinez2023improving} implicitly account for interactions between agents by learning strategies through interacting with the environment. However, they lack formal guarantees, are vulnerable to out-of-distribution data, require significant training time, and produce a non-parameterized policy that cannot be easily adjusted if it fails to meet user requirements.
Alternatively, MPC-based planners~\cite{de2023globally, martinez2024shine, de2024topology} explicitly account for deterministic~\cite{brito2019model} or uncertain~\cite{zhu2019chance, wang2020fast} future behaviors of humans when planning trajectories.

Despite significant advancements over the past decade, developing effective and reliable navigation solutions for robots in human-centered environments remains a complex challenge. 
Effective collision avoidance alone is not enough to ensure human acceptance; addressing aspects such as comfort, naturalness, and sociability in navigation is also crucial~\cite{kruse2013human}. 
Many of the above approaches require defining desired behaviors through a cost or reward function, which demands a deep understanding of both the system and the application environment, and do not facilitate on-the-fly behavior adaptation.

\textbf{LLMs in Robotics.} In the last years, the application of LLMs and VLMs has revealed new opportunities for enhancing autonomous systems.
VLMs hold significant potential for mobile manipulation, in particularly in determining the sequence of steps needed to complete tasks~\cite{huang2022language,liu2024enhancing,jin2024robotgpt,winge2024talk}. These models have been employed not only to generate the necessary code for planning and executing these intermediate sub-tasks \cite{liang2023code} but also to enable real-time, interactive communication with the robot guiding it \cite{lynch2023interactive}. Differently from others, \cite{kwon2024language} presents a trajectory generation algorithm for the end-effector of a manipulator understanding language prompts with an LLM, enabling motion planning. 

In mobile robots, trajectory generation from VLMs is also used to achieve zero-shot object navigation~\cite{majumdar2022zson,dai2024think}.
In autonomous driving, for example, there are methods that incorporate them to understand the environment and output a desired car trajectory~\cite{sima2023drivelm}, directly control commands~\cite{shao2024lmdrive, chen2024driving} or select one command from a list~\cite{sha2023languagempc}. However, the methods do not have formal guarantees and have not proved to work in a close loop in real time. As in previous works, we aim to use VLMs for high-level understanding and interactive behavior. Unlike them, we introduce the VLM inside an MPC formulation to modify the robot behavior in real time.

\textbf{Personalized robot navigation.} Recent works incorporate the gathering of human feedback to modify robot trajectory generation, and use language model to process them. \cite{bucker2022reshaping,bucker2023latte} use a network to reshape trajectories given feedback and a previous trajectory, while \cite{Sharma-RSS-22} generates a costmap from a human query to be included and combined with other costs for trajectory optimization. These approaches require previous training and are for manipulators in static environments. They can not be applied in our problem, which needs fast control feedback for scenario changes. 

More related to our work, some works incorporate physical interaction~\cite{kollmitz2020learning} and virtual reality~\cite{de2022learning,de2023learning} feedback in a reinforcement learning socially-aware motion planner for mobile robots. Nevertheless, our approach is distinct from others in that it is zero shot, uses language and visual information and has collision avoidance constraints.

\section{Problem formulation}

\begin{figure*}
    \centering
    \includegraphics[width=\textwidth]{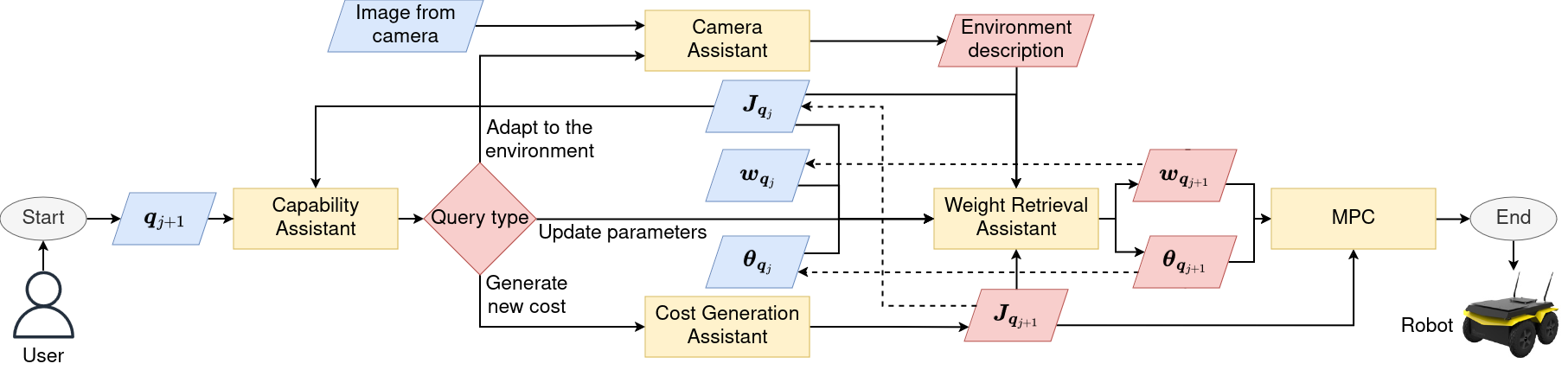}
    \caption{Flow diagram of our proposed LLM module and its connection to the MPC. Inputs are in blue and outputs in red. Dashed lines represent updates regarding the state for the following query.}
    \label{fig:llm_nav}
\end{figure*}

We consider a robot with discrete time non-linear dynamics:
\begin{equation}
    \boldsymbol{x}_{k+1}=f(\boldsymbol{x}_k, \boldsymbol{u}_k),
\end{equation}
\noindent where $\boldsymbol{x}_k \in \mathbb{X} \subseteq \mathbb{R}^{N_x}$ is the robot state and $\boldsymbol{u}_k \in \mathbb{U} \subseteq \mathbb{R}^{N_u}$ the control input at time $k$. The state contains robot position, $\boldsymbol{p}_k=\{x_k,y_k\} \in \mathbb{R}^2$, and  radius $r_r$. As the environment is dynamic, we consider $\mathcal{X}:=\mathbb{R}^2\times[0,T]$ as the robot's workspace, with $[0,T]$ a continuous finite time domain.

The robot has to navigate in an environment with moving obstacles and avoid collisions with them and the rest of obstacles. We assume that the moving obstacles are humans, but could be other robots. The humans are represented as disks with radius $r_h$. The position of human $i$ at time step $k$ is $\boldsymbol{o}_{i,k} \in \mathbb{R}^2$, the position of all humans is denoted $\mathcal{O}_k$ and the area occupied by static obstacles in the environment is denoted $\mathcal{S} \subseteq \mathbb{R}^2$. The robot knows its position, the positions of the surrounding humans, their predicted trajectory and the local costmap of the scenario. It is also equipped with a camera to capture images of its environment.

As the robot navigates, the user can provide instructions in natural language, referred to as \emph{queries}. We denote a query $j$ as $\bm{q}_j$, where $\bm{q}_{j+1}$ is the query received after $\bm{q}_{j}$. Our goal is to navigate as specified by query $\bm{q}_j$ and, with less emphasis, the queries $\bm{q}_0, \hdots, \bm{q}_{j-1}$ as long as they do not contradict $\bm{q}_j$. 
We formalize the problem as a nonlinear optimization problem over the time horizon $N_k$:%
\begin{subequations}\label{eq:mpc_problem}%
\begin{align}%
    \min_{\boldsymbol{x}\in\mathbb{X}, \boldsymbol{u}\in\mathbb{U}} \quad & \sum_{k=0}^{N_k} J_{\bm{q}_j}(\boldsymbol{x}_k, \boldsymbol{u}_k, \boldsymbol{w}_{\bm{q}_j}, \boldsymbol{\theta}_{\bm{q}_j}) \label{eq:J}\\
    \textrm{s.t.} \quad \quad & \boldsymbol{x}_{k+1}=f(\boldsymbol{x}_k, \boldsymbol{u}_k) \ \forall k, \label{eq:dynamics-sub}\\
    & \boldsymbol{x}_0 = \boldsymbol{x}_{init}, \label{eq:dynamics-init-sub}\\
    & g(\boldsymbol{x_k},\boldsymbol{\theta}_k)\leq 0 \ \forall k, \label{eq:restrictions-sub}
\end{align}
\end{subequations}
where in Eq.~\ref{eq:J} the cost function that is generated to satisfy $\bm{q}_j$ is denoted $J_{\bm{q}_j}$. 
The values $\boldsymbol{w}_{\bm{q}_j}\in\mathbb{R}^{N_{w_{\boldsymbol{q}_j}}}$ weight the terms in $J_{\bm{q}_j}$ (e.g., $w_\alpha$ weights $\mathcal{J}_\alpha$), and $\boldsymbol{\theta}_{\bm{q}_j}$ is a set of parameters related to the robot behavior. These include tunable parameters such as a reference velocity and environment-related parameters such as $\mathcal{O}_k$, $\mathcal{S}$, a user specified goal, $\bm{p}_{goal} \in \mathbb{R}^2$, or a reference path. 
Eq.~\ref{eq:dynamics-sub} denotes the robot's dynamic constraints, Eq.~\ref{eq:dynamics-init-sub} the initial conditions and Eq.~\ref{eq:restrictions-sub} the collision avoidance constraints. For 
simplicity, we model the robot dynamics with a second-order unicycle model in this paper, but the method may be applied for any robot dynamics. We consider as control input $\boldsymbol{u}_k=(a_k, \omega_k)$ linear acceleration and angular velocity, respectively.

\section{Approach}
In this section, we propose Hey Robot!, a novel planning architecture (see Fig.~\ref{fig:llm_nav}) that uses LLMs to automatically generate and tune the cost function in Eq.~\ref{eq:mpc_problem}, optimized by an MPC, in order to generate safe trajectories that comply with user instructions. 
Its MPC produces control commands every control step and is reconfigured on the fly by the LLM components
when a new query, $\bm{q}_{j+1}$, is received.

\subsection{LLM components}

The system has four different LLM components, which we call assistants, in charge of different tasks. 
All assistants are queried to give their answers as short as possible to limit interaction time. We do ask for a brief motivation to allow the LLM to reason, improving accuracy. 
In this section, we use the query $\bm{q}_j:$ "Follow the path" as a running example to clarify our approach. To that end, let the cost $J_{\bm{q}_j}$ generated with respect to $\bm{q}_j$ include the following components: contour, $\mathcal{J}_c$, and lag, $\mathcal{J}_l$, terms to track a reference path in $\boldsymbol{\theta}_{q_j}$ as defined in~\cite{brito2019model}; input penalty cost terms to penalize each of the inputs $\mathcal{J}_a=a^2$, $\mathcal{J}_\omega=\omega^2$; and a velocity tracking cost, $\mathcal{J}_v=(v_k-v_{ref})^2$, to track a reference velocity, $v_{ref}$ in $\boldsymbol{\theta}_{q_j}$: $J_{\bm{q}_j}=J_{path}=w_c\mathcal{J}_c + w_l\mathcal{J}_l + w_v\mathcal{J}_v + w_a\mathcal{J}_a + w_\omega\mathcal{J}_\omega$. 
Our method can handle queries specifying behavior (e.g., ``drive faster'') and task assignments (e.g., ``go to the goal'').
\subsubsection{Capability Assistant}

When the system receives a new prompt $\boldsymbol{q}_{j+1}$, it first needs to decide if the current cost function, $J_{\boldsymbol{q}_j}$, is sufficient. We consider that $J_{\boldsymbol{q}_j}$ is sufficient for $\boldsymbol{q}_{j+1}$ if $\boldsymbol{q}_{j+1}$ can be achieved by minimizing $J_{\boldsymbol{q}_j}$ or any cost term in $J_{\boldsymbol{q}_j}$. 
This assistant uses the LLM to understand $\boldsymbol{q}_{j+1}$ and $J_{\boldsymbol{q}_j}$, and forwards the query to the next most suitable assistant out of three possible options:
\begin{itemize}
    \item \textbf{Generate a new cost}, if $J_{\boldsymbol{q}_j}$ is not sufficient for $\boldsymbol{q}_{j+1}$. This implies that $\boldsymbol{q}_{j+1}$ refers to a different task like "reach the goal" or "follow a human". Then, the next assistant is the \textit{Cost Generation Assistant}.
    \item \textbf{Adapt to the environment}, if $\boldsymbol{q}_{j+1}$ asks the robot to sense and adapt to its surrounding environment without any additional information (e.g. "Perceive the scenario"). The query is forwarded to \textit{Camera Assistant}.
    \item \textbf{Update parameters}, if $J_{\boldsymbol{q}_j}$ is sufficient for $\boldsymbol{q}_{j + 1}$. In our example, $\boldsymbol{q}_{j+1}$ could include behavioral changes like "be faster" or "be smoother". In that case, $J_{\boldsymbol{q}_{j+1}} = J_{\boldsymbol{q}_j}$ and the next assistant is the \textit{Weight Retrieval Assistant}.
\end{itemize}

\subsubsection{Cost Generation Assistant}
The goal of this assistant is to generate a new cost function $J_{\boldsymbol{q}_{j+1}}$ that satisfies $\boldsymbol{q}_{j+1}$. 
Our approach asks the LLM to freely design $J_{\boldsymbol{q}_{j+1}}$ given the query, directly writing the code of the cost function. Cost terms in $J_{\boldsymbol{q}_{j+1}}$ may depend on $\boldsymbol{x}_k$, $\boldsymbol{u}_k$ and parameters $\bm{\theta}_{\bm{q}_{j+1}}$. Notice that through $\bm{\theta}_{\bm{q}_{j+1}}$ the LLM can devise new parameters that could be updated in later queries. 

To limit generation errors, the structure of the code function to be generated is firstly specified. Additionally, a list of typical cost terms is provided for two reasons. First, we allow the LLM to name these cost terms instead of writing their code, reducing response times. Second, so that it may use them as code example to generate others (few-shot prompting). The list of provided terms is $\mathcal{J}_c$, $\mathcal{J}_l$, $\mathcal{J}_a$, $\mathcal{J}_\omega$, $\mathcal{J}_v$ and a cost to reach the goal position, $\mathcal{J}_g=||\boldsymbol{p}_{goal}-\boldsymbol{p}_k||_2^2$. We command the LLM to always include $\mathcal{J}_{a}$, $\mathcal{J}_\omega$ and $\mathcal{J}_v$ as part of the cost function. If it needs to generate new cost terms different than the predefined ones, we query it to use a quadratic cost to minimize, its inverse to maximize and \texttt{if\_else} CasADi function to check conditions enforcing it to use smooth and differentiable operations. 

In our example, if $\boldsymbol{q}_{j+1}$ was "Reach the goal", the new cost function should be $J_{\boldsymbol{q}_{j+1}}=J_{goal}=w_g\mathcal{J}_g + w_v\mathcal{J}_v + w_a\mathcal{J}_a + w_\omega\mathcal{J}_\omega$. Nevertheless, if $\boldsymbol{q}_{j+1}$ was "Follow the closest human", then a human following term should be generated from scratch, resulting in $J_{\boldsymbol{q}_{j+1}}=J_{hf} = w_{hf}||\boldsymbol{o}_{h,k}-\boldsymbol{p}_k||_2^2 + w_v\mathcal{J}_v + w_a\mathcal{J}_a + w_\omega\mathcal{J}_\omega$, where the closest human $\boldsymbol{o}_{h,k} \!= \! \argmin_{\boldsymbol{o}_{i,k} \in \mathcal{O}_k} ||\boldsymbol{o}_{i,k}-\boldsymbol{p}_k||_2^2$ is retrieved using CasADi \texttt{if\_else}.

The LLM generated cost function $J_{\boldsymbol{q}_{j+1}}$ is used to regenerate solver binaries and the solver is then reloaded in the MPC to activate it. 
The values of $\boldsymbol{\theta}_{\boldsymbol{q}_{j+1}}$ and $\boldsymbol{w}_{\boldsymbol{q}_{j+1}}$ are tuned with the \textit{Weight Retrieval Assistant}.

\subsubsection{Camera Assistant}
This assistant uses the camera to perceive the environment and understand its implications for navigation. It allows the robot to assess for example if its environment is crowded, empty, narrow or open, and can recognize specific environments such as a hospital or warehouse environment. 
First, a photo is taken from a robot onboard camera and is passed to the VLM that is requested to output a list of bullet points related to ideal robot motion in the perceived scenario. We set $J_{\boldsymbol{q}_{j+1}} = J_{\boldsymbol{q}_j}$ to ensure that this information is solely used to inform the navigation behavior, not to change task. The description of the environment is then forwarded to the \textit{Weight Retrieval Assistant}.


\subsubsection{Weight Retrieval Assistant}

This assistant assigns $\boldsymbol{w}_{\boldsymbol{q}_{j+1}}$ and tunable parameters in $\boldsymbol{\theta}_{\boldsymbol{q}_{j+1}}$. Its LLM prompt includes the code of $J_{\boldsymbol{q}_{j+1}}$ and the values of $\boldsymbol{w}_{\boldsymbol{q}_{j}}$ and $\boldsymbol{\theta}_{\boldsymbol{q}_{j}}$. It also includes as query $\boldsymbol{q}_{j+1}$ or the extracted environment description when triggered by the \textit{Camera Assistant}.

As LLMs are known to struggle with numerical tasks, it is difficult to reliably tune weights $\boldsymbol{w}_{\boldsymbol{q}_{j+1}}$ with an LLM. To circumvent this issue we instead ask the LLM to rate the importance of each cost term in $J_{\bm{q}_{j+1}}$ as an integer between $0$ and $10$, jointly denoted as $\boldsymbol{z}_{\boldsymbol{q}_{j+1}}$. 
In our example, the query "stick to the path" would result in $z_c$ and $z_l$ close to 10, while "be smoother" would increase the previous value of $z_a$ and $z_\omega$. 
We then convert each importance rating $z_{\alpha}$ into a weight $w_{\alpha}$ using 
$w_\alpha=\frac{z_\alpha}{\overline{\boldsymbol{z}}_{\boldsymbol{q}_{j+1}}}$, where $\overline{\boldsymbol{z}}_{\boldsymbol{q}_{j+1}}$ is the mean value of the elements in $\boldsymbol{z}_{\boldsymbol{q}_{j+1}}$. 
For the tunable parameters in $\boldsymbol{\theta}_{\boldsymbol{q}_{j+1}}$, the LLM is, in our experience, able to generate reasonable values as these parameters are absolute and have a physical meaning that the LLM understands. We therefore query the LLM directly for their values. 
For example, the query "be faster" would increase $v_{ref}$.

\subsection{MPC formulation}

The MPC controls the robot in closed loop solving the optimization in Eq.~\ref{eq:J} every control period. We use CasADi~\cite{Andersson2018casadi} and Acados~\cite{Verschueren2021acados} in the implementation. Our proposed formulation is similar to the one in \cite{de2023globally, de2024topology}. Having $J_{\boldsymbol{q}_j}$, the collision constraints in Eq.~\ref{eq:restrictions-sub} include human avoidance:
\begin{equation}
    g(\boldsymbol{x}_k, \boldsymbol{o}_{j,k}) = 1-\Delta\boldsymbol{p}_{j,k}^T\bm{R}_k^T \begin{pmatrix}
    \frac{1}{r^2} & 0 \\
    0 & \frac{1}{r^2}
    \end{pmatrix}
    \bm{R}_k\Delta\boldsymbol{p}_{j,k},
\end{equation}
\noindent
where $\Delta\boldsymbol{p}_{j,k}=\boldsymbol{p}_k-\boldsymbol{o}_{j,k}$ and $r=r_r+r_h$. We also include static collision avoidance with $\mathcal{S}$, using the \textit{Safe Flight Corridor} concept, as in \cite{liu2017planning} or \cite{benders2024embedded}. 

MPC generally optimizes the robot trajectory from an initial guess, leading to a locally optimal trajectory that avoids the obstacles based on that guess. As the environment changes, local optimality can lead to poor and potentially dangerous navigation behavior. To compute a global optimal trajectory and avoid local minima, we use the approach from~\cite{de2024topology}. Several guidance trajectories from different homology classes~\cite{bhattacharya2012topological} are sampled and optimized in parallel using Acados, resulting locally optimal. We select the one with the minimum cost to control the robot.
\section{Results}

We conducted experiments to test the assistant capabilities and the navigation system's behavior, both in simulation and on a ground robot. We used GPT-4o-mini~\cite{achiam2023gpt} as the LLM via its public API. The implementation used ROS with modules in C++ and Python. The MPC cost code uses Python to avoid recompilation after new cost generation reducing processing time. The assistants are evaluated in Sec.~\ref{sec:assis-exp} and the navigation system in Sec.~\ref{sec:nav-exp}.


\subsection{Assistants experiments}\label{sec:assis-exp}

We assessed the assistants individual performance with a battery of specific queries for each, stated in Table~\ref{tab:assistants-prompts}. We tried each prompt 10 times and gathered the results.

\begin{table}
    \caption{Prompts tested in each experiment.}
    \centering
    \begin{tabular}{c|l}
        \textbf{Code} & \textbf{Prompt} \\
        \hline
        C1 & Go to the goal. You are navigating through a hospital. \\
        C2 & Stick to the path. \\
        C3 & Follow the closest human. \\
        C4 & Go to the goal while keeping a safe distance from humans. \\
        C5 & Adapt to the environment. \\
        \hline        
        G1 & Follow the path. \\
        G2 & Reach the goal. \\
        G3 & Maximize the distance to the closest human. \\
        G4 & Minimize the distance to the closest human. \\
        G5 & Go to the goal while keeping a safe distance from humans. \\
        G6 & Follow the closest human. \\
        \hline
        W1 & Be faster. \\
        W2 & Take more distance to humans. \\
        W3 & Stick to the path. \\
        W4 & Be smoother. \\
        W5 & Increase rotation capabilities. \\
        W6 & You can rotate more. \\
    \end{tabular}
    \label{tab:assistants-prompts}
\end{table}

\subsubsection{Capability assistant}

\begin{figure}[t]
    \centering
    \includegraphics[width=\linewidth]{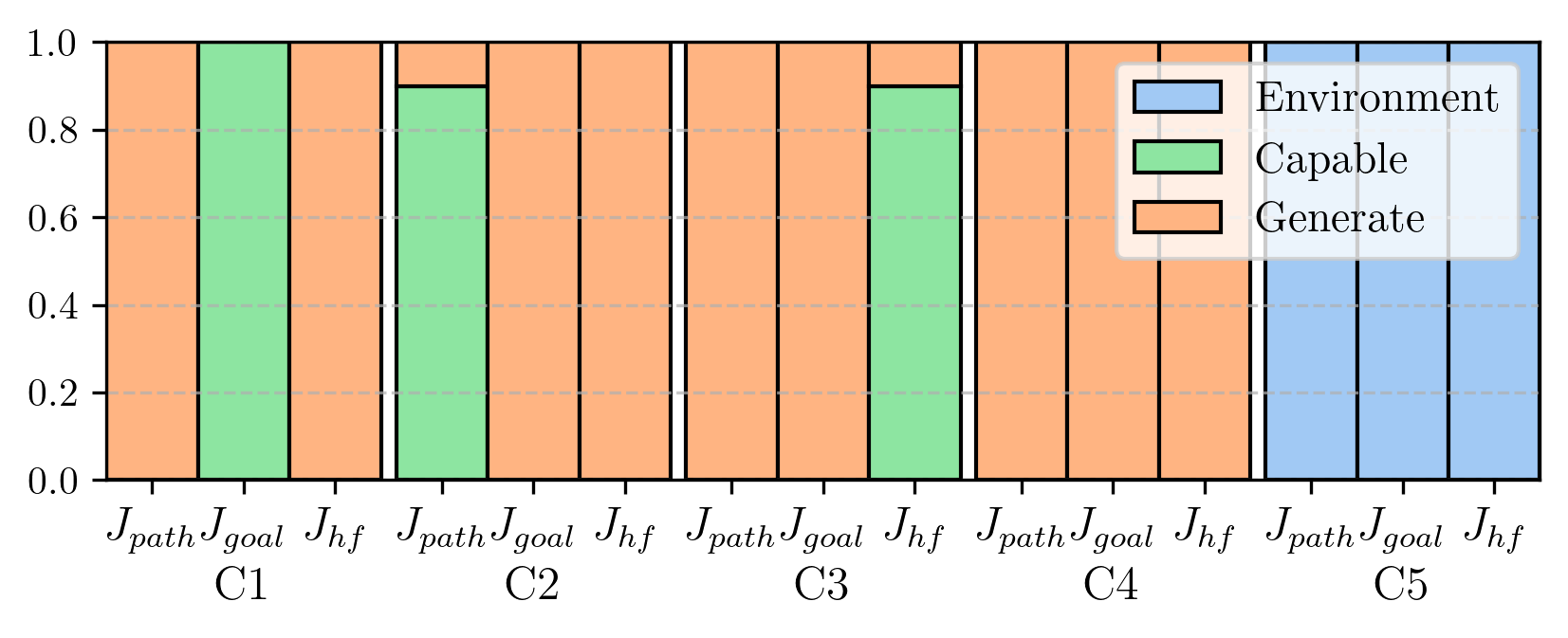}
    \caption{Number of times (rate) each response was selected for the \textit{Capability Assistant} experiment.}
    \label{fig:capability-success}
\end{figure}

The experiment uses queries in Table~\ref{tab:assistants-prompts} as $\boldsymbol{q}_{j+1}$ and $J_{path}$, $J_{goal}$ and $J_{hf}$ as $J_{\boldsymbol{q}_j}$. Fig.~\ref{fig:capability-success} represents the response rate for each $\boldsymbol{q}_{j+1}$ and $J_{\boldsymbol{q}_j}$ for different colors for the three possible responses. The expected results are that $J_{path}$ satisfies C2,  $J_{goal}$ satisfies C1, $J_{hf}$ satisfies C3, none of them C4 and it should adapt to the environment with C5. As the assistant has direct access to the code, written and commented using the same LLM, it successfully completes its task most of the time. The assistant only fails twice, being over-conservative, deciding to generate a new cost when it is not needed. This incurs a larger generation time, but the system still performs its task successfully.

\subsubsection{Cost Generation Assistant}

For this assistant, we verify that it reliably generates a cost function that fulfills $\boldsymbol{q}_{j+1}$ without syntax or runtime errors. We test:
\begin{itemize}
    \item G1 and G2: Differentiating between path and goal following. In this case, it may use the names of a predefined cost terms provided in list in the prompt.
    \item G3 to G6: Designing by itself different cost terms that were not provided. It should design terms that maximize (G3) or minimize (G4) variables, or design high-level primitives like "keep a safe distance" (G5) or "follow a human" (G6). 
\end{itemize}

Table~\ref{tab:gen-weight-succ} shows the success rates of the cost generation assistant. The assistant generates $J_{path}$ to follow a path (G1) and $J_{goal}$ to reach a goal (G2). It generates new cost terms that minimize (G3, G6) and maximize (G4) the distance to $\boldsymbol{o}_{h,k}$ as needed, using $J_{hf}$ and $J_{hmax} = w_{hmax}\frac{1}{(\boldsymbol{o}_{h,k}-\boldsymbol{p}_k)^2+\epsilon} + w_v\mathcal{J}_v + w_a\mathcal{J}_a + w_\omega\mathcal{J}_\omega$, respectively. Finally, it can also create a floating point parameter like the safe distance in G5 that may be tuned without generating a new cost, generating $J_{sd} = w_{sd}\mathcal{J}_{sd} + w_g\mathcal{J}_g + w_v\mathcal{J}_v + w_a\mathcal{J}_a + w_\omega\mathcal{J}_\omega$, having that $d_{safe}\in\boldsymbol{\theta}_{sd}$ is the safe distance; and $\mathcal{J}_{sd} = (||\boldsymbol{o}_{h,k}-\boldsymbol{p}_k||_2^2-d_{safe})^2$ is only applied if $||\boldsymbol{o}_{h,k}-\boldsymbol{p}_k||_2^2-d_{safe} > 0$ using \texttt{if\_else} CasADi function. The failure case in G3 was produced because it minimized instead of maximized the distance to the closest human once (we hypothesize that it is because G3 is not a common query), and in G5 because it hard coded the safe distant as a constant once, instead of including it as a parameter.


\begin{table}[t]
    \caption{\textit{Cost Generation} and \textit{Weight Retrieval} assistants success rates.}
    \centering
    \begin{tabular}{c|cccccc}
         \textbf{Assistant}&  \textbf{G1} & \textbf{G2} & \textbf{G3} & \textbf{G4} & \textbf{G5} & \textbf{G6}\\
         \textit{Cost Generation} & 1.0 & 1.0 & 0.9 & 1.0 & 0.9 & 1.0 \\
         \hline
         \textbf{Assistant}&  \textbf{W1} & \textbf{W2} & \textbf{W3} & \textbf{W4} & \textbf{W5} & \textbf{W6}\\
         \textit{Weight Retrieval} & 1.0 & 1.0 & 1.0 & 0.8 & 1.0 & 1.0 \\
    \end{tabular}
    \label{tab:gen-weight-succ}
\end{table}

\subsubsection{Weight Retrieval Assistant}

For these experiments, we use $J_{sd}$ to test W2 and $J_{path}$ to test the rest of the queries. The success rates of the weight retrieval assistant are shown in Table~\ref{tab:gen-weight-succ}, computed by checking if the corresponding weights and parameters tuning correspond to the query intentions. We check the final relative values of the weights to measure the results. The results show that the assistant tunes the parameters in $\boldsymbol{\theta}_{q_j}$ as expected, increasing $v_{ref}$ and $d_{safe}$ for W1 and W2, respectively. It also increases $w_v$ and $w_{sd}$ with those queries, which is desirable. With W3 to W6, it proves that it understands high-level queries. It increases $w_c$ and $w_l$ in W3, increases $w_a$ and $w_\omega$ in W4 and reduces $w_\omega$ in W5 and W6. The failure cases in W4 are because, as well as increasing the $z_a$ and $z_\omega$, it increases more $z_c$ and $z_l$, so final $w_a$ and $w_\omega$ do not increase. However, in deployment, this could be simply solved by trying the same query again or being more specific.


\begin{figure}[t]
    \centering
    \captionsetup{justification=centering}
    \begin{subfigure}[t]{0.31\linewidth}
        \centering
        \includegraphics[width=\linewidth]{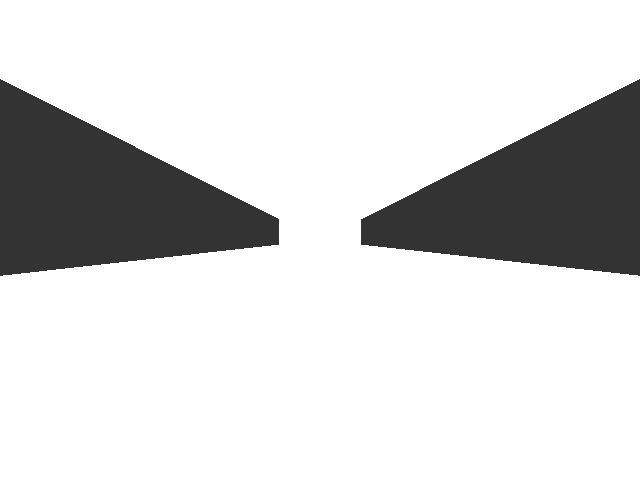}
        \subcaption{}
        \label{fig:corridor}
    \end{subfigure}\hspace{5pt}%
    \begin{subfigure}[t]{0.31\linewidth}
        \centering
        \includegraphics[width=\linewidth]{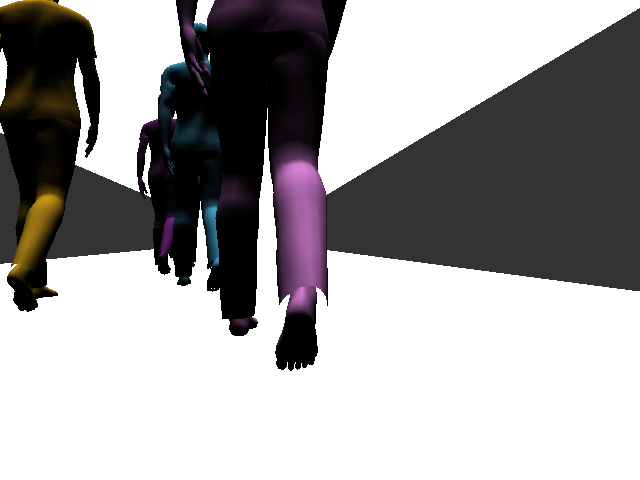}
        \subcaption{}
        \label{fig:crowded_corridor}
    \end{subfigure}\hspace{5pt}%
    \begin{subfigure}[t]{0.31\linewidth}
        \centering
        \includegraphics[width=\linewidth]{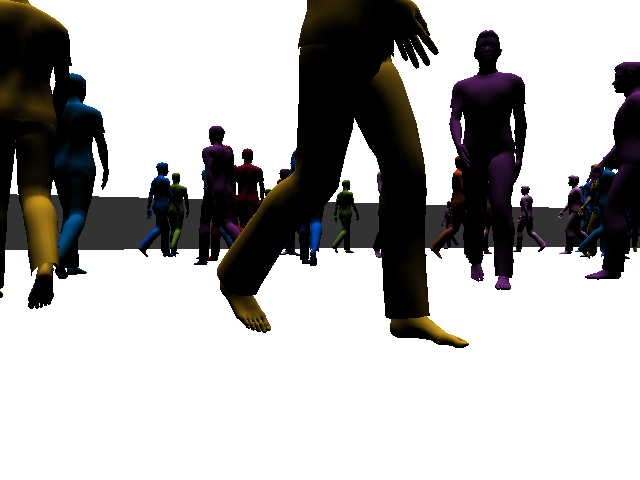}
        \subcaption{}
        \label{fig:crowded_open}
    \end{subfigure}
    \caption{Images of the simulator taken with the robot camera.}
    \label{fig:camera-assistant}
\end{figure}

\subsubsection{Camera assistant}

In this experiment, we qualitatively test the \textit{Camera Assistant} with the \textit{Weight Retrieval Assistant} in a simulated scenario. We use $J_{path}$ as $J_{\bm{q}_{j+1}}$. We set all initial values in $\boldsymbol{z}_{q_{j}}$ to 5, and use the images in Fig.~\ref{fig:camera-assistant} for the queries. The results, shown in Table~\ref{tab:camera-assistant}, are the following:

\begin{enumerate}
\renewcommand{\labelenumi}{\alph{enumi})}
    \item The scenario is described as a confined scenario with no dynamic obstacles. Thus, it sets as the most important weights the ones that track the reference path and reference velocity.
    \item The scenario is described as a narrow pathway with human congestion, with special requirement in careful navigation. It sets high the path tracking cost, due to the environment being narrow, but reduces substantially the velocity tracking weight, as being smooth and successful is more important around humans.
    \item The camera detects multiple humans in close proximity and a high crowd density. As a confined room is not detected, path tracking weight are set lower, and the most important aspect is set to be a smooth navigation.
\end{enumerate}

\begin{table}[t]
    \centering
    \caption{Final $\boldsymbol{z}_{q_{j+1}}$ given the images in Fig.~\ref{fig:camera-assistant}.}
    \begin{tabular}{c|cccccc}
    \textbf{Image} & $z_c$ & $z_l$ & $z_v$ & $z_a$ & $z_\omega$\\
    \hline
        (a) & 8 & 8 & 6 & 6 & 7  \\
        (b) & 8 & 8 & 4 & 7 & 6  \\
        (c) & 6 & 6 & 4 & 7 & 7 \\
    \end{tabular}

    \label{tab:camera-assistant}
\end{table}

\begin{table}
    \centering
    \caption{Mean and standard deviation of times spent in query processing.}
    \begin{tabular}{c|cccc}
      \textbf{Assistant}   & Capability & Cost Gen. & Weight Ret. & Camera\\
      \hline
      \textbf{Time (s)} & 1.91 (1.04) & 4.02 (1.63) & 1.16 (0.23) & 1.79 (0.38)
    \end{tabular}
    \label{tab:assistants-time}
\end{table}

\begin{table*}
\small\centering
\caption{Mean and standard deviation of 10 simulations for each of six queries in the corridor-like scenario of Sec.~\ref{sec:sim_queries}.}\resizebox{\textwidth}{!}{%
\begin{tabular}{l|ccccccc}
\textbf{Additional Instructions} & \textbf{Col. rate} & \textbf{Dur. [s]} & \textbf{Path len [m]} & \textbf{Min Dist. [m]} & \textbf{$\bm{v}$ [m/s]} & \textbf{$\bm{a}$ [m/$\bm{\text{s}^2}$]} & \textbf{$\bm{\omega}$ [rad/s]} \\
\hline
a) (Default) & \textbf{0.00} & 14.1 (0.8) & 34.99 (0.32) & 0.22 (0.04) & 2.49 (0.12) & 0.51 (0.09) & 0.18 (0.04) \\
b) Drive quickly. & \textbf{0.00} & \textbf{13.7} (1.1) & 34.77 (0.42) & 0.19 (0.06) & \textbf{2.56} (0.17) & 0.59 (0.14) & 0.18 (0.05) \\
c) Drive carefully. & \textbf{0.00} & 24.1 (0.7) & 33.30 (0.20) & 0.15 (0.04) & \textbf{1.38} (0.04) & 0.13 (0.05) & \textbf{0.05} (0.02) \\
d) You are navigating through a factory without humans. & \textbf{0.00} & 13.8 (0.8) & 35.03 (0.57) & 0.33 (0.05) & 2.55 (0.12) & 0.56 (0.09) & 0.20 (0.07) \\
e) You are navigating through a hospital. & \textbf{0.00} & 23.6 (0.9) & 33.28 (0.11) & 0.25 (0.06) & 1.41 (0.05) & \textbf{0.10} (0.05) & \textbf{0.05} (0.02) \\
f) Try to keep a distance of at least 1.5m from pedestrians. & \textbf{0.00} & 19.8 (5.6) & \textbf{31.09} (6.75) & \textbf{1.16} (0.20) & 1.74 (0.59) & 0.38 (0.17) & 0.16 (0.06) \\
\end{tabular}}
\label{tab:comparison}
\end{table*}

\subsubsection{Processing times}

The times taken to process the queries by the LLM in the assistant experiments presented in this section is shown in Table~\ref{tab:assistants-time}. The time to process the queries takes only a few seconds, which makes real-time interaction with the user possible. Regarding updates that do not require cost generation, the user may see the changes in the robot in around $3$s (\textit{Capability Assistant} and \textit{Weight Retrieval Assistant} times), while with queries that may change completely the behavior of the robot only around $7$s (adding \textit{Cost Generation Assistant} time).

\subsection{Navigation Results} \label{sec:nav-exp}

We tested our complete navigation stack with all its components working together with different queries, evaluating generated trajectories and metrics. 

\subsubsection{Task simulations}

\begin{figure}[t]
    \centering
    \begin{subfigure}[t]{0.49\linewidth}
        \centering
        \includegraphics[width=\linewidth]{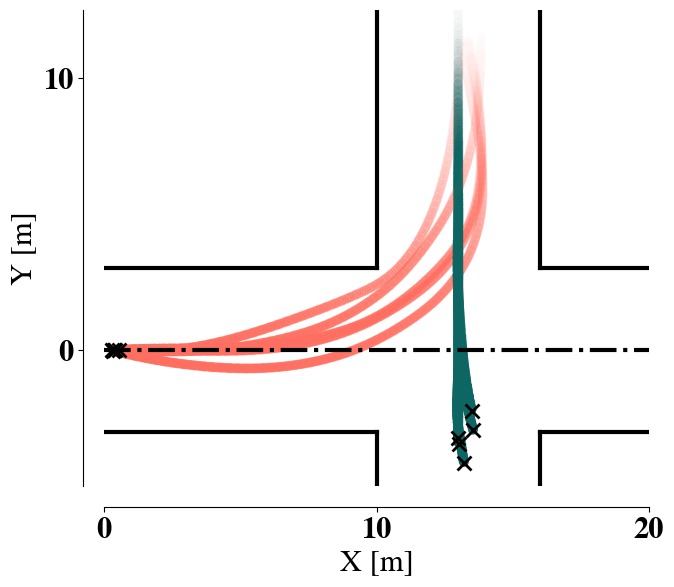}
        \subcaption{$\boldsymbol{q}_j$: Follow the closest human.}
        \label{fig:trajectories_path}
    \end{subfigure}%
    \hspace{1pt}
    \begin{subfigure}[t]{0.49\linewidth}
        \centering
        \includegraphics[width=\linewidth]{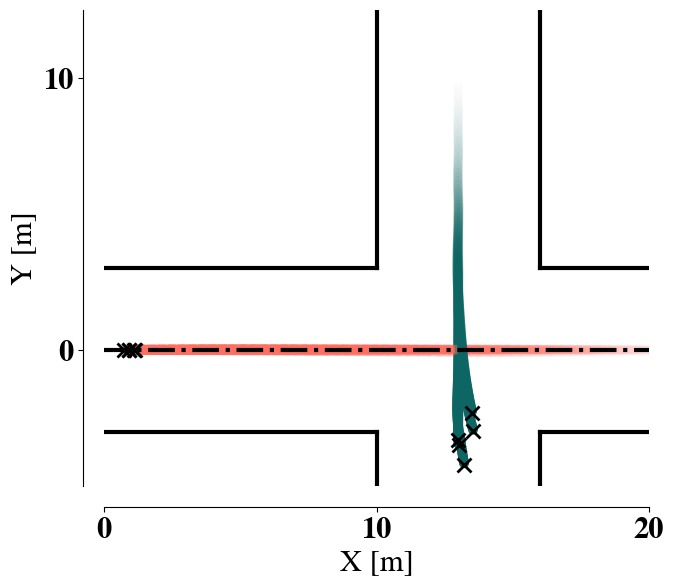}
        \subcaption{$\boldsymbol{q}_j$: Follow the path.}
        \label{fig:trajectories_pedestrian}
    \end{subfigure}
    \caption{Five overlapping experiments with trajectories of the robot (blue) and human (green) and a reference path (dashed).}
    \label{fig:trajs-path-ped}
\end{figure}

First, we tested our method in with a simulated Clearpath Jackal robot and pedestrians moving using Social Forces~\cite{helbing1995social}. We used a constant velocity assumption to predict the future trajectories of the humans. We first verify if our method allows the user to change the robot's task. We performed five simulations for the two queries: ``follow the reference path'' and ``follow the closest human''. The results in Fig.~\ref{fig:trajs-path-ped} indicate that two \emph{distinct} behaviors emerge, each completing the user's task. The trajectories differ across simulations due to variations in $\boldsymbol{w}_j$ and $\boldsymbol{\theta}_j$, as the queries do not specify additional information.


\subsubsection{Navigation metrics}\label{sec:sim_queries}

We conducted ten simulations for each of six different user queries in a corridor-like scenario, with a reference path going through the center of the corridor. All queries included the task "Follow the reference path" (default), plus additional instructions. Table~\ref{tab:comparison} shows the results. None of the experiments lead to collisions, as the MPC has collision avoidance restrictions. It may be observed that the fastest and the ones with highest velocities are b) and d), as they are told to be fast and productive. c) and e), on the contrary have lower velocities and are smoother, using smaller input commands. f) has the highest minimum distance to the closest pedestrian metric, as stated in the instructions, but its velocity and episode duration is not low, as other instructions where not provided. It also compromises penalizing inputs and time to reach the goal, rather than the others which prefer one of them. Overall, the navigation metrics match the high-level intention of the queries.

\subsubsection{Laboratory experiments}

\begin{figure}[t]
    \centering
    \begin{subfigure}[t]{0.47\linewidth}%
        \centering
        \includegraphics[trim={0 0 10.9cm 0},clip,width=\textwidth]{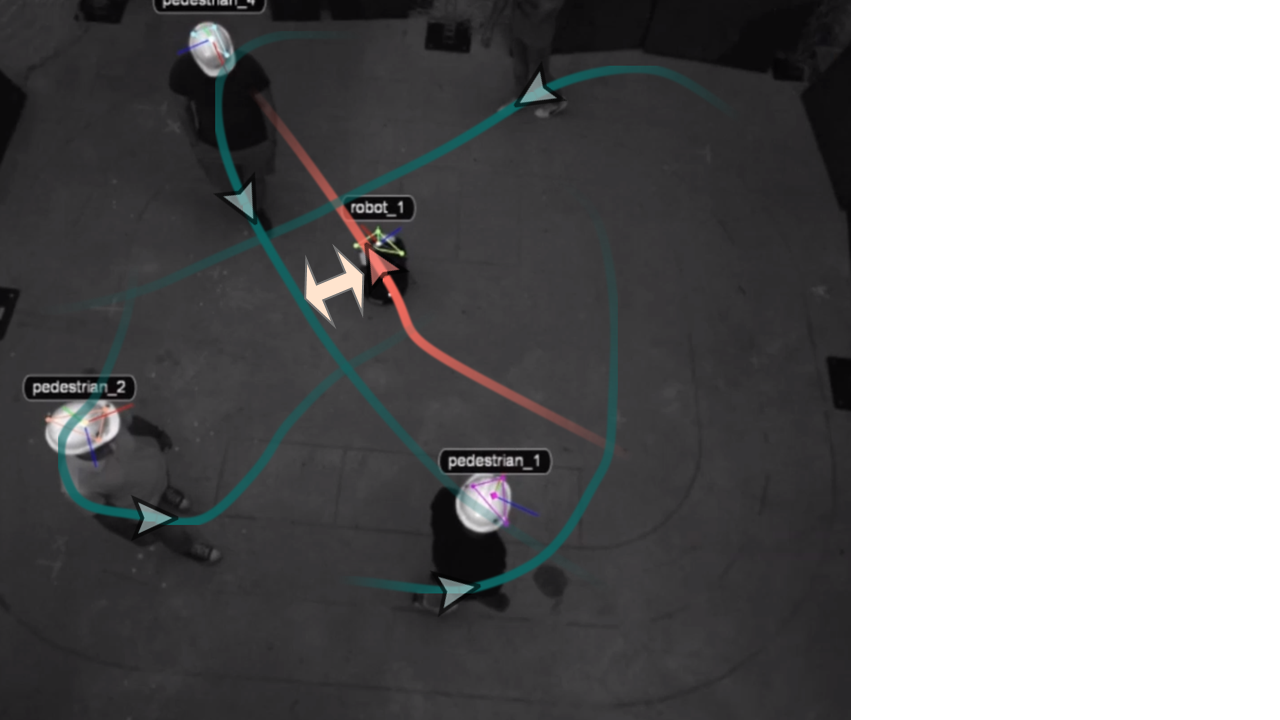}
        \caption{$\boldsymbol{q}_j$: Go to the goal.}
    \end{subfigure}%
    \hspace{1.0pt}%
    \begin{subfigure}[t]{0.47\linewidth}%
        \centering
        \includegraphics[trim={0 0 10.9cm 0},clip,width=\textwidth]{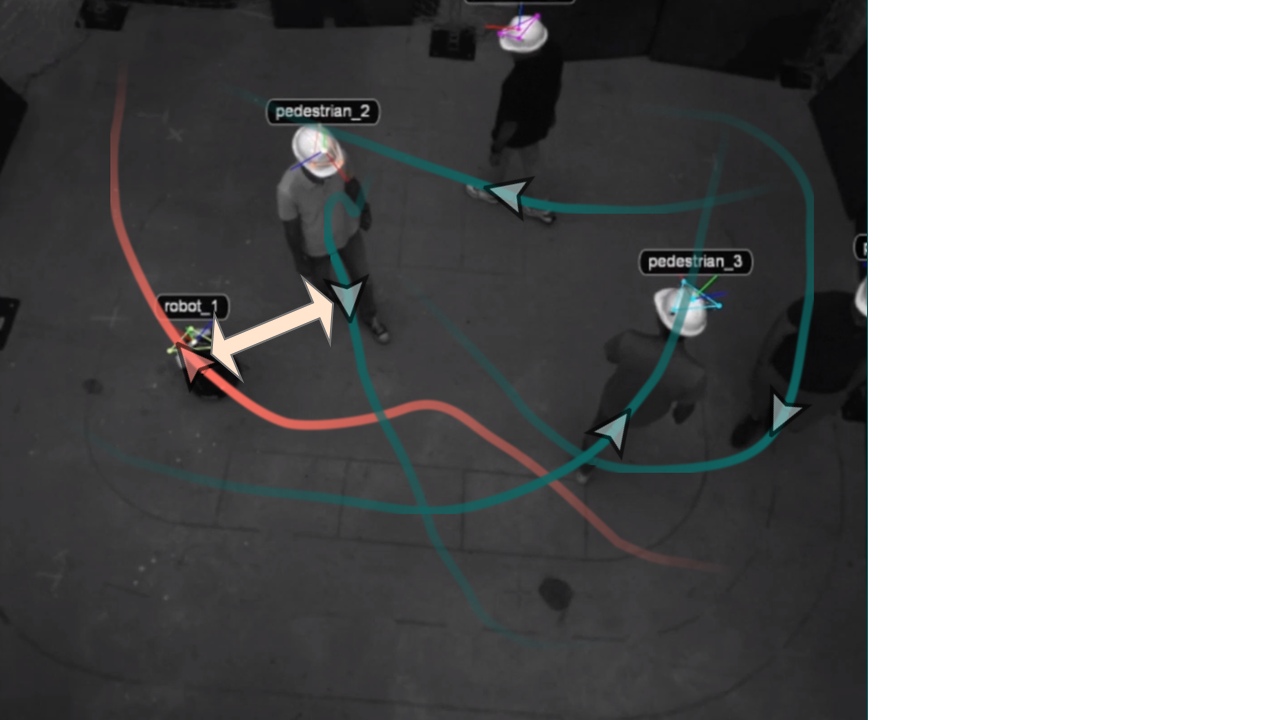}
        \caption{$\boldsymbol{q}_j$: Go to the goal while keeping $1$m from humans.}
    \end{subfigure}%
    \caption{Robot navigating with four humans.}
    \label{fig:lab-exp}
\end{figure}

Moreover, we tested the ability of our method to change the robot behavior on the fly. We used a Turtlebot 2 as the ground platform, an Optitrack system with twelve cameras for localization and a Kalman Filter with constant velocity assumption to estimate the velocity of four pedestrians present. The experiment consists in a user constantly giving different instructions to the robot while it navigates. The robot effectively addresses the user queries in a successful navigation, proving the effectiveness of our solution, as seen in Fig.~\ref{fig:lab-exp}. Simulation and real-world videos of our method changing the robot behavior on the fly may be seen in the supplementary video.

\section{Conclusion}

This work presented a novel approach for personalizing mobile robot navigation behavior on the fly with natural language. We leveraged LLMs to understand the human reasoning behind the user queries, and used the information to generate and tune the cost function of an MPC designed for safe navigation in dynamic environments. Our experiments showed the system's successful performance in simulations and in a real robot, demonstrating navigation behavior that matched the user requirements. With this work, we pave the way for new methods that reduce the gap between robots and regular users. As LLMs improve, the processing time and query understanding of our system should improve. This work is limited to the cost function inputs and perception. Further work will include scene understanding and segmentation to support object interaction, and experimentation in realistic environments with different users.

\section*{Acknowledgements}
The use of GPT-4 public API is integrated as part of the proposed method, as explained in the manuscript.





\bibliographystyle{IEEEtran}
\bibliography{references}
\balance
\end{document}